\documentclass[conference]{IEEEtran}
\IEEEoverridecommandlockouts
\usepackage{multirow}
\usepackage{graphicx}
\usepackage{amsmath,amssymb,amsfonts}
\usepackage{booktabs}
\usepackage{algorithm,algorithmic}
\usepackage[algo2e,linesnumbered]{algorithm2e} 
\usepackage{pifont}  
\usepackage{subfigure}
\usepackage{graphicx}
\usepackage{textcomp}
\usepackage{xcolor}
\usepackage{mathtools}
\usepackage{bbm}
\usepackage{soul}
\usepackage{cite}
\usepackage{makecell}
\usepackage[linkcolor=black,
            anchorcolor=black,
            citecolor=black]{hyperref}
\def\BibTeX{{\rm B\kern-.05em{\sc i\kern-.025em b}\kern-.08em
T\kern-.1667em\lower.7ex\hbox{E}\kern-.125emX}}
\begin{document}

\title{Spatio-Temporal Pruning for Compressed Spiking Large Language Models}

 \author{
\IEEEauthorblockN{Yi Jiang\IEEEauthorrefmark{1}, Malyaban Bal\IEEEauthorrefmark{1}, Brian Matejek\IEEEauthorrefmark{2}, Susmit Jha\IEEEauthorrefmark{2}, Adam Cobb\IEEEauthorrefmark{2}, Abhronil Sengupta\IEEEauthorrefmark{1}}
\IEEEauthorblockA{\IEEEauthorrefmark{1}School of Electrical Engineering and Computer Science, The Pennsylvania State University, University Park, PA 16802, USA \\
Email: \{yijiang, mjb7906, sengupta\}@psu.edu}
\IEEEauthorblockA{\IEEEauthorrefmark{2}SRI International, Arlington, USA \\
Email: \{brian.matejek, susmit.jha, adam.cobb\}@sri.com}
}

\maketitle


\begin{abstract}

Large Language Models present significant challenges for deployment in energy-constrained environments due to their large model sizes and high inference latency. Spiking Neural Networks, inspired by the sparse event-driven neural processing and energy-efficient information transmission in the brain, offer a promising alternative for achieving low-power computing. Integrating the event-driven efficiency of spiking neurons with the advanced capabilities of LLMs represents a promising direction for power-efficient LLMs. This work specifically delves into the design of compressed spiking LLMs. 
Here, we revisit spatial and temporal pruning from the perspective of SNNs and propose a novel spatio-temporal pruning framework for Spiking LLMs to optimize computational efficiency while preserving high performance. Our spatial pruning technique reduces the number of active neurons and attention heads, effectively lowering the computational complexity of the model. Meanwhile, temporal pruning minimizes inference latency by dynamically adjusting the number of timesteps required for different layers. We present the first work in the domain of Spiking LLMs that introduces a two-stage spatio-temporal pruning framework, where spatial and temporal redundancies are systematically pruned, together with extreme quantization and knowledge distillation strategies.
Extensive experimental evaluation of our proposed framework for SpikingBERT on the large-scale GLUE benchmark demonstrates the efficacy of our approach in terms of computational operations and inference latency. Our approach offers a compelling solution for real-time, low-power natural language processing applications, making Spiking LLMs more practical for deployment on edge devices and in power-constrained settings.
\end{abstract}

\begin{IEEEkeywords}
Large Language Models, Spiking Neural Networks, Pruning, Quantization, Knowledge Distillation.
\end{IEEEkeywords}

\section{Introduction}
Large Language Models (LLMs) have revolutionized the field of natural language processing, providing remarkable advances in tasks such as text generation and machine translation. LLMs, like GPT \cite{radford2018improving} and BERT \cite{devlin2019bertpretrainingdeepbidirectional}, have set new benchmarks in terms of accuracy and performance, demonstrating the vast potential of deep learning in understanding and generating human language. However, the large number of parameters and high computational demands translate to considerable energy consumption, limiting their feasibility for real-time and low-power applications.  To address these challenges, several compression methods have been developed, including quantization \cite{kim2021ibertintegeronlybertquantization, MLSYS2024_42a452cb}, knowledge distillation (KD) \cite{jiao2020tinybertdistillingbertnatural, sanh2020distilbertdistilledversionbert}, pruning \cite{kwon2022fastposttrainingpruningframework, NEURIPS2023_44956951, Wang_2021}, low-rank factorization \cite{hu2021loralowrankadaptationlarge, ben-noach-goldberg-2020-compressing}, and other techniques. These methods aim to reduce the model size of LLMs without significantly compromising their performance. From a complementary compute perspective, Spiking Neural Networks (SNNs) \cite{MAASS19971659}, which mimic the spike-based information processing found in biological neural systems, have emerged as a promising alternative for resource-efficient computing \cite{10.3389/fnins.2019.00095}. The potential of SNNs has been demonstrated in several practical real-world applications \cite{10878462, 11125950}. Integrating LLMs with SNNs presents a compelling opportunity to utilize the strengths of both architectures, combining the high accuracy and language understanding capabilities of LLMs with the power efficiency and real-time processing capability of SNNs. Recent works showcase the benefits of using spiking architectures in LLMs, including SpikingBERT \cite{bal2024spikingbert} and SpikeLLM \cite{xing2024spikellmscalingspikingneural}. Meanwhile, advances in spiking Transformer architectures, such as Spikformer \cite{zhou2022spikformerspikingneuralnetwork}, SpikingViT \cite{10586833}, and SSTFormer \cite{wang2025sstformer}, further highlight the development of spike-driven Transformer models.

Although spike-based LLMs can reduce power consumption through the inherent sparsity of spiking neurons, the model still requires a substantial amount of operations and retains a large number of parameters. As a result, the overall computational cost still remains high, limiting the scalability and efficiency of spike-based LLMs in practical applications. Further, unlike traditional neural networks, which process information in a single forward pass, SNNs require multiple timesteps to achieve comparable accuracy. \textit{Hence there is an unmet need to explore compression techniques catered for spiking LLMs that consider their unique spiking and temporal properties.}

Compression techniques (like quantization, pruning, and knowledge distillation), specifically designed for SNNs, have been explored in Refs. \cite{kushawaha2020distillingspikesknowledgedistillation, xing2024spikellmscalingspikingneural,QIAO2021203,bal2024exploringextremequantizationspiking, chowdhury2021spatiotemporalpruningquantizationlowlatency, bal2024spikingbert}. 
Pruning, in particular, has proven to be an effective strategy for reducing both the model size and inference latency of spiking neurons \cite{10.1145/3477145.3477157,rathi2017stdpbasedpruningconnections}. In non-spiking models, pruning typically removes redundant or less critical neurons and attention heads, streamlining the network architecture and significantly reducing the number of active parameters. Unlike non-spiking networks, pruning in SNNs must also consider the event-driven and temporal nature of spiking activity, where reducing spiking events and simulation timesteps directly lead to lower energy consumption and latency. As a result, temporal pruning can be introduced to further optimize SNNs \cite{chowdhury2021spatiotemporalpruningquantizationlowlatency,10758407}. 
However, existing temporal pruning methods primarily reduce simulation length by uniformly shortening timesteps during training. In contrast, our temporal pruning formulation treats timesteps as layer-wise allocatable computational resources and is realized through a stochastic rate-based propagation scheme, which not only avoids storing full spike trajectories but also enables efficient and flexible temporal optimization while preserving effective temporal information flow. \textit{Moreover, the vast majority of prior works remain limited to simple vision-based datasets and convolutional architectures. Although initial work \cite{bal2024spikingbert, bal2024exploringextremequantizationspiking, xing2024spikellmscalingspikingneural} has explored quantization and knowledge distillation in spiking LLMs, pruning remains relatively understudied, especially in the context of developing a cohesive framework combining various compression techniques.}
\\
\noindent The specific contributions of our work are as follows:\\
\noindent \textbf{(i) Revisiting Spatio-Temporal Pruning in the Spiking LLM Setting:} We rethink and extend conventional spatial and temporal pruning strategies by adapting them specifically for SNNs. We propose a principled two-stage pruning framework for Spiking LLMs that incorporates both post-training and retraining phases. In the post-training phase, spatial pruning is guided by a combination of Fisher Information Matrix (FIM), average spiking rate (ASR), and attention output to identify and remove redundant neurons and attention heads. Concurrently, temporal pruning is performed using Principal Component Analysis (PCA) to dynamically allocate timesteps based on the importance of each layer. Furthermore, we integrate PCA analysis for temporal pruning and a computational cost penalty into the loss function for spatial pruning during the retraining phase to enforce energy-aware fine-tuning. This two-stage design ensures that the model achieves high computational efficiency without compromising task performance. 
\textit{Notably, this is the first work to jointly explore the combination of spatial and temporal pruning for large-scale spiking LLMs.} 
This unified pruning strategy is specifically designed to exploit the spatial and temporal characteristics of spiking LLMs.

\noindent \textbf{(ii) Performance Evaluation of Compressed Spiking LLMs on the GLUE Benchmark:} \textit{We present the first training framework integrating spatial pruning, temporal pruning, extreme quantization, and knowledge distillation for Spiking LLMs evaluated on the large-scale GLUE benchmark}. Specifically, our experiments focus on SpikingBERT as a representative spiking language model. Prior SNN compression works have primarily focused on small-scale datasets, limiting their applicability. Our method demonstrates strong scalability and practical feasibility across multiple GLUE tasks, making it a more viable and efficient solution for deploying Spiking LLMs in real-world, energy-constrained environments.

\section{Related Works}
\noindent \textbf{Spiking LLMs:} SNNs have emerged as a promising approach to develop compute-efficient LLMs. 
Spikeformer \cite{zhou2022spikformerspikingneuralnetwork} is the first work to use the spike-based self-attention mechanism, marking a significant step in this direction. SpikeGPT \cite{zhu2024spikegptgenerativepretrainedlanguage} provides the first demonstration of language generation enabled by direct training of SNNs. SpikingBERT \cite{bal2024spikingbert} uses the average spiking rate of neurons at equilibrium to train the model using implicit differentiation techniques. By avoiding the need for surrogate gradients or explicit time unrolling during model training, this approach enables more efficient and scalable training, providing a solid foundation for further enhancements such as quantization and pruning. We therefore choose SpikingBERT training framework as our baseline for exploring spatio-temporal pruning strategies.

\noindent \textbf{Compression of Spiking LLMs:} 
Compression techniques for Spiking LLMs remain in the nascent stage, with works primarily targeting quantization and knowledge distillation. Refs. \cite{xing2024spikellmscalingspikingneural, bal2024exploringextremequantizationspiking} investigated quantization methods for Spiking LLMs to reduce weight precision. Knowledge distillation methods, explored in Refs. \cite{lv2024spikebertlanguagespikformerlearned, bal2024spikingbert}, train smaller Spiking LLMs using larger pre-trained models as teachers. This work primarily focuses on developing a cohesive training framework for Spiking LLMs that incorporates pruning strategies along with extreme quantization and knowledge distillation. 

\noindent \textbf{Pruning Methods:} 
Pruning methods in the ANN domain are generally categorized into unstructured and structured pruning. Unstructured pruning removes individual weights and yields sparse parameter matrices, as exemplified by PLATON \cite{zhang2022platon}. Structured pruning removes entire structural components, such as neurons or attention heads, resulting in more hardware-friendly architectures. EBERT \cite{liu2021ebert} explores structured pruning by dynamically removing attention heads and feed-forward channels during inference, while CoFi \cite{xia2022structured} formulates structured pruning as a learnable process that identifies redundant attention heads and intermediate neurons.
More recently, PGB \cite{lim2025pgb} proposes a one-shot semi-structured pruning strategy that bridges unstructured and structured pruning. Given a trained model, pruning typically proceeds in two stages: removing less important model components, followed by fine-tuning to recover performance. Most works focus on the post-training stage \cite{Wang_2021, zhang2024plugandplay, Lazarevich_2021_ICCV},  as it avoids the substantial cost of training the large-scale models from scratch. To guide the pruning process more effectively, some approaches leverage the Hessian matrix of the loss function with respect to the model parameters, primarily in the context of unstructured pruning. Specifically, the weights associated with smaller eigenvalues of the Hessian have less impact on the model’s performance and can be pruned with minimal loss. The technique has been explored and proven effective in prior works \cite{NIPS1989_6c9882bb, 298572} for efficiently pruning less critical weights. However, the computational complexity of these methods has limited their scalability to larger models. To address this limitation,  Refs. \cite{kwon2022fastposttrainingpruningframework, liu2021groupfisherpruningpractical} utilized FIM \cite{Fisher1925TheoryOS} as an approximation of the Hessian to perform post-training structured pruning. The FIM is data-aware, computationally efficient, and empirically effective, enabling importance estimation without explicit Hessian construction. These properties make it particularly well-suited for post-training pruning of large-scale transformer architectures. In this work, we adopt this Fisher-based approach and extend it with additional techniques tailored for spiking language models, incorporating spiking-specific metrics, such as average spiking rate, to guide the pruning process. 

\noindent \textbf{Pruning Methods for SNNs:} Pruning algorithms have been explored in SNN literature. EQ-SpikeLM \cite{11152565} applies structured spatial pruning by removing redundant attention heads and feed-forward components, and further combines post-training quantization to construct a lightweight spiking language model. Ref. \cite{Chen_2021} proposed a gradient-based rewiring pruning method to adjust the network structure by rewiring connections dynamically. In Refs. \cite{rathi2017stdpbasedpruningconnections, Neftci_2016}, weight connections are pruned based on the membrane potential. Refs. \cite{10.3389/fnins.2020.598876, 10.3389/fnins.2019.00405} proposed adaptive weight pruning for SNNs. Ref. \cite{shi2024towards} presented a method for an unstructured pruning framework with energy consumption constraints. Ref. \cite{chowdhury2021spatiotemporalpruningquantizationlowlatency}  explored the use of PCA to achieve structured pruning for SNNs, a method also used in ANNs \cite{Garg_2020}. These works all focus on spatial pruning in SNNs. \textcolor{black}{Other studies have also explored alternative approaches to improving computational efficiency in SNNs \cite{11120352}.} Unlike ANNs, SNNs have an additional critical dimension: the temporal dimension, which also plays an important role in their efficiency and performance \textcolor{black}{\cite{11316659}}. Ref. \cite{chowdhury2021spatiotemporalpruningquantizationlowlatency} proposed a pruning framework focusing on reducing temporal redundancies during the training phase. This approach starts with an SNN trained over multiple timesteps and iteratively reduces the number of timesteps uniformly across the entire network. A similar strategy is also proposed in \cite{inbook}.  However, this uniform approach disregards the varying importance of different layers, which could potentially benefit from a more layer-specific temporal pruning strategy. Ref. \cite{kim2022exploringtemporalinformationdynamics} investigated the temporal information dynamics in SNNs, revealing that shallow layers exhibit higher importance during the later timesteps, while deeper layers demonstrate significant importance in earlier timesteps. This insight provides the motivation to explore layer-specific temporal pruning. Ref. \cite{10888589} proposed a dynamic timestep approach that utilizes layer-wise binary masks to reduce inference latency and energy consumption. However, when an early layer is masked, its output becomes zero, which prevents later layers from receiving any input. This can significantly degrade the performance of deeper layers, which are typically more critical for final decision-making. As a result, the effectiveness of this approach is limited, particularly in deep architectures or more complex tasks, and has so far only been demonstrated on small-scale vision benchmarks. This motivates us to develop a more flexible and scalable approach that allows each layer to allocate its own number of timesteps independently, without being constrained by earlier layers. In the domain of spatio-temporal pruning, Ref. \cite{10758407} integrates both spatial and temporal pruning for Spiking Transformers. However, similar to prior discussions,  temporal pruning is applied uniformly across all layers without adaptation to layer-specific dynamics, and the spatial pruning relies solely on spike accumulation values, limiting scalability and flexibility for more complex tasks or architectures.
\textit{In summary, the vast majority of pruning works explored for SNNs remain limited to small scale tasks and simple convolutional architectures, with only limited works targeting spatio-temporal pruning. Our work overcomes these critical challenges to develop a unified spatio-temporal pruning strategy for Spiking LLMs equipped with quantization and knowledge distillation.}

\begin{figure*}[h]
    \centering
    \includegraphics[ scale=0.68, trim=0 1.2cm 0cm 0,clip
    ]{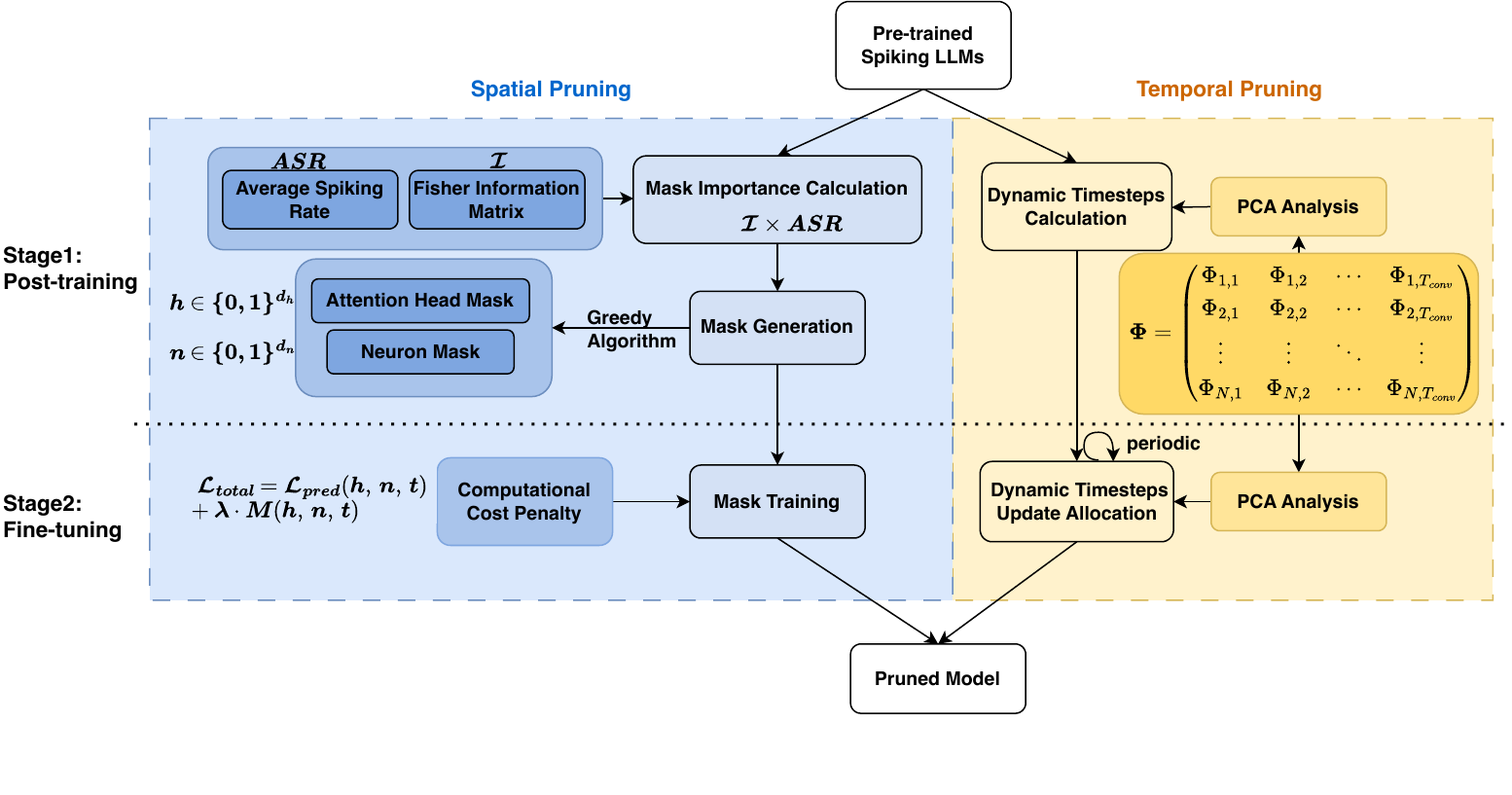}
    \caption{Overview of the two-stage pruning framework for Spiking LLMs. In Stage 1 (Post-training stage), spatial pruning is performed by calculating importance scores based on the FIM and ASR, which are used to generate neuron and attention head masks. Temporal pruning is guided by a PCA-based analysis of a layer-wise spiking activity matrix $\boldsymbol{\Phi}$, which captures the temporal spiking dynamics across layers and timesteps, to determine initial timestep allocations for each layer. The resulting masks and timestep configurations serve as initialization for the Stage 2 (fine-tuning stage). During fine-tuning, the model is trained over multiple epochs, where masks are updated with a computational cost penalty $M(h,n,t)$ to promote sparsity, and layerwise timestep allocations are updated every few epochs by using PCA analysis. This framework enables energy-efficient and low-latency inference in pruned Spiking LLMs.}
    \label{fig:framework}
\end{figure*}
\section{Methods}

In this section, we first introduce the spiking neuron model in SNNs and the associated training technique considered in this work. Subsequently, we present a comprehensive two-stage pruning methodology designed specifically for Spiking LLMs. As shown in Fig.~\ref{fig:framework}, our approach integrates both spatial and temporal pruning techniques to enhance computational efficiency and energy savings, while maintaining model performance. The first stage focuses on post-training pruning. For spatial pruning, we utilize the FIM, ASR, and attention output to identify the redundant neurons and attention heads. This accelerates the model by reducing the number of required operations and enhances energy efficiency by reducing the number of spikes processed during inference. Since the output of each SNN layer is binary spikes instead of continuous activation values, computation involves only accumulation operations (ACs). Therefore, we use the number of ACs to quantify computational cost. For temporal pruning, we apply PCA analysis to dynamically adjust the number of timesteps across layers, reducing temporal redundancy. In the second stage, we refine the model via fine-tuning with pruning. For spatial pruning, we add a penalty term based on ACs to the loss function to reduce overall computational load. For temporal pruning, we periodically perform PCA analysis to dynamically reallocate timesteps across layers. Further details on justification for using FIM, ASR, and attention output during spatial pruning and PCA during temporal pruning will be provided in later sections.

For the training process, we adopt the same approach used in SpikingBERT\cite{bal2024spikingbert} (see Appendix \ref{arch} for details on architecture), which leverages implicit differentiation to compute gradients at a steady-state convergence of the ASR of neurons rather than using Backpropagation Through Time (BPTT). This approach significantly reduces the computational overhead and memory requirements. Additionally, the steady-state ASR convergence of neurons in linear as well as non-linear layers of the  SpikingBERT model enables the usage of knowledge distillation, where a pre-trained BERT model serves as the teacher to guide the training of the spiking student model. Quantization of model weights can also be easily integrated into the training framework \cite{bal2024exploringextremequantizationspiking}.

\subsection{SpikingBERT Training Framework}

We first introduce the discrete time dynamics of an LIF-based spiking neuron, which is the fundamental component of our SpikingBERT framework,
\begin{equation}
\label{eqn1}
u_i^t = \gamma  u_i^{t-1} +  \sum_j ({w_{i,j}s_j^{t-1}})+ b_i - s_i^{t-1}*V_{th}, 
\end{equation}
\begin{equation}
s_i^t = \hat{\Theta}(u_i^t, V_{th})
\end{equation}
where, $u_i^t$ is the membrane potential of neuron $i$ at time $t$, $w_{i,j}$ is the weight connecting the pre-synapse neuron $j$ and
post-synapse neuron $i$, $s_j^{t-1}$ denotes the input spiking signal of pre-
synapse neuron $j$ at time $t-1$,  $b_i$ indicates a bias term, $\gamma$ is the leaky constant and $V_{th}$ is the spiking threshold. $\hat{\Theta}$ is the non-differentiable spiking operation with an underlying Heaviside function, which is defined as,
\begin{equation}
\begin{aligned}
\hat{\Theta}(u_i^t, V_{th}) = H(u_i^t -  V_{th})
\end{aligned}
\end{equation}
where, ${H(x) = \begin{cases} 
0 & \text{if } x < 0 \\
1 & \text{if } x \ge 0 
\end{cases} }$. Subtraction serves as the reset operation for $u$ in neurons that spiked.

During training, SpikingBERT leverages implicit differentiation at equilibrium \cite{xiao2021trainingfeedbackspikingneural}, to overcome the non-differentiability of spike-based dynamics. This approach computes gradients based solely on the final converged steady-state of ASR, which is reached after $T_{conv}$ timesteps of convergence. The gradient of the loss $\mathcal{L}_{pred}$ with respect to parameters $\theta$ is computed as:

\begin{equation}
\label{eqn5}
\frac{\partial \mathcal{L}_{pred}(a^*)}{\partial \theta} = - \frac{\partial \mathcal{L}_{pred}(a^*)}{\partial a^*} (J^{-1}_{g_{\theta}}|_{a^*}) \frac{\partial f_{\theta}(a^*)}{\partial \theta}
\end{equation}
where, $a$ denote the ASR value, $a^*$ is the steady-state ASR value, $f$ is the steady-state equation of ASR, $g_{\theta}(a) = f_{\theta}(a) - a$, $J^{-1}_{g_{\theta}}$ is the inverse Jacobian of $g_\theta$ when $a = a^{*}$, i.e., at equilibrium. Instead of relying on BPTT, which requires storing the entire sequence of intermediate states, this approach avoids temporal unrolling by computing gradients at equilibrium, thereby significantly reducing memory consumption and enabling scalable training.

\subsection{Post-training Pruning}\label{POST}
\subsubsection{Spatial Pruning} To explicitly control the computational cost of the pruned model, we introduce a constraint on the total number of ACs, which reflects the energy and latency requirements of spiking architectures. The detailed computation of ACs is provided in Appendix \ref{ACs}. Since only spatial pruning is considered in this case, all layers operate with the same number of timesteps. Therefore, the timestep variable $t$ can be omitted from the ACs calculation (see Appendix~\ref{ACs}). In the post-training spatial pruning stage, we determine the optimal binary masks for pruning while satisfying this ACs constraint. We employ two types of masks: neuron masks and head masks. The neuron mask is applied in the feedforward layers to selectively prune less significant neurons, and the head mask is used in the multi-head attention layers to prune redundant attention heads. The pruning problem is formulated as follows:
\begin{equation}
\label{argmin}
\arg\min_{h, n} \mathcal{L}_{pred}(h, n) \quad \text{s.t.} \quad M(h, n) \leq M_c
\end{equation}
where, $\mathcal{L}_{pred}$ is the prediction loss function, $h \in \{0,1\}^{d_h}$ is the head mask, and $d_h$ is the number of attention heads. Similarly, $n \in \{0,1\}^{d_n}$ is the neuron mask, where $d_n$ is the number of intermediate neurons. $M$ represents the ACs of the pruned network, and $M_c$ is the target ACs constraint.  By constraining the total number of ACs, we ensure that the pruned network remains within a desired energy budget. The selection criteria for $M_c$ is further explored in our ablation studies (see Section \ref{experiement:ablation}). 
 
We approximate the prediction loss around the unpruned model $(h=\mathbf{1}, n=\mathbf{1})$ by using the second-order Taylor expansion as:

\begin{equation}
\begin{aligned}
\mathcal{L}_{\mathrm{pred}}(h,n) \approx\;&
\mathcal{L}_{\mathrm{pred}}(\mathbf{1},\mathbf{1}) -
 g^{\!\top} \begin{bmatrix}
\mathbf{1}-h\\
\mathbf{1}-n
\end{bmatrix}
\\
&+\frac{1}{2}
\begin{bmatrix}
\mathbf{1}-h\\
\mathbf{1}-n
\end{bmatrix}^{\!\top}
H
\begin{bmatrix}
\mathbf{1}-h\\
\mathbf{1}-n
\end{bmatrix}
\end{aligned}
\label{eq:taylor_expansion}
\end{equation}
where, $H$ denotes the Hessian matrix of the prediction loss with respect to the pruning masks, and $g$ denotes the gradient of the prediction loss with respect to the pruning masks, defined as:
\begin{equation}
\begin{aligned}
g &=
\left.
\nabla_{(h,n)}
\mathcal{L}_{\mathrm{pred}}(h,n)
\right|_{h=\mathbf{1},\,n=\mathbf{1}}, \\
H &=
\left.
\nabla^2_{(h,n)}
\mathcal{L}_{\mathrm{pred}}(h,n)
\right|_{h=\mathbf{1},\,n=\mathbf{1}}
\end{aligned}
\end{equation}
Following previous pruning methods \cite{kwon2022fastposttrainingpruningframework}, we use the formulation in Eqn. \ref{eq:taylor_expansion} as a heuristic approximation to characterize the sensitivity of the prediction loss to the pruning masks, rather than as a rigorous second-order estimate of the pruning-induced loss change. Accordingly, the first-order term is not explicitly included in the final optimization objective.

Furthermore, $\mathcal{L}_{\mathrm{pred}}(\mathbf{1},\mathbf{1})$ is a constant, it can be omitted from the optimization objective. Therefore, the pruning problem in Eqn. \ref{argmin} can be rewritten as:
\begin{equation}
\arg\min_{h,n}
\begin{bmatrix}
\mathbf{1}-h \\
\mathbf{1}-n
\end{bmatrix}^{\!\top}
H
\begin{bmatrix}
\mathbf{1}-h \\
\mathbf{1}-n
\end{bmatrix}
\quad \text{s.t.} \quad M(h,n) \leq M_c
\label{eq:hessian_form}
\end{equation}

However, solving this constrained optimization problem directly is intractable due to the binary, non-differentiable nature of the mask variables. To address this issue, we approximate the Hessian using a diagonal empirical FIM with respect to the mask variables. The FIM provides a second-order approximation of the loss increase when a parameter is removed, thereby quantifying its importance to the model. Thus, instead of minimizing the loss directly, we minimize the cumulative Fisher scores of the pruned units. This leads to the following final objective:
\begin{equation}
\label{objective}
\arg \min_{h,n} \left( \sum_{i \in S(h)} \mathcal{I}_{ii}+ \sum_{j \in S(n)} \mathcal{I}_{jj} \right)
\end{equation}
where,  $S(h) := \{i \mid {h}_i = 0\}$ and $S(n) := \{j \mid {n}_j = 0\}$, and $\mathcal{I}$ is FIM of gradient of masks,  which is defined as:
\begin{equation}
\begin{aligned}
\mathcal{I}_{ii}
&=
\mathbb{E}_{(x,y)\sim\mathcal{X}}
\left[
\left(
\frac{\partial \mathcal{L}_{{pred}}(x,y)}{\partial h_i}
\right)^2
\right], \\
\mathcal{I}_{jj}
&=
\mathbb{E}_{(x,y)\sim\mathcal{X}}
\left[
\left(
\frac{\partial \mathcal{L}_{{pred}}(x,y)}{\partial n_j}
\right)^2
\right]
\end{aligned}
\end{equation}
where, $\mathcal{X}$ denotes the training sample set used to estimate the empirical Fisher information. $(x,y)$ denotes an input–label pair sampled from the training set, and the expectation is approximated using empirical averages over training samples. The diagonal elements of the FIM represent the individual sensitivity of each mask with respect to the loss function. In our setting, we treat each of these elements as an importance score: a larger diagonal value indicates that the corresponding unit has a greater influence on the loss, while a smaller value suggests it can be safely pruned. This allows us to rank neurons and attention heads based on their estimated contribution to model performance and selectively retain only the most critical ones. 

\textit{Unlike conventional neural networks, SNNs communicate using discrete spikes, resulting in inherently sparse and temporally dynamic activity. Motivated by this spiking-specific characteristic, we design a post-training pruning framework that combines traditional sensitivity-based measures, like the FIM, with spike-driven activity statistics, such as ASR and layer-wise timestep usage.} Intuitively, an effective spatial pruning criterion for spiking LLMs should satisfy two requirements:
(i) it should capture how sensitive the prediction loss is to removing a unit, and (ii) it should reflect how frequently the unit participates in spiking computation, which directly impacts energy consumption. Many existing pruning criteria primarily emphasize only one of these aspects. Our method explicitly combines both factors to guide pruning decisions in a principled and spiking-aware manner. The ASR is a direct measure of how frequently each neuron contributes to the computation, making it an essential factor in determining neurons' importance. 
Therefore, the final importance score for each neuron mask should be a combination of both the FIM and the corresponding ASR. Similarly, for the attention mechanism, a head with larger output values should have a more substantial impact on the outputs of the fully connected layer, as well as the entire transformer block. We use the ASR of the attention outputs to assess the importance of each head, reflecting how frequently a head contributes to computation. This idea conceptually aligns with prior work in non-spiking models, where head importance is estimated by accumulating the absolute values of attention outputs, as shown in Ref. \cite{Wang_2021}. 
The final importance of neuron masks and head masks can be summarized as follows:
\begin{equation}
I(h_i) = \mathcal{I}_{ii} \cdot  ASR_i,
\end{equation}
\begin{equation}
I(n_j) = \mathcal{I}_{jj} \cdot ASR_j
\end{equation}
where, $\mathcal{I}_{ii}$ and $\mathcal{I}_{jj}$ are the Fisher information importance scores for head $i$ and neuron $j$ respectively. $ASR_i$ and $ASR_j$ denote the average spiking rate of the attention output of head $i$ and the average spiking rate of neuron $j$, respectively. This design jointly considers prediction sensitivity and spiking activity instead of relying on either metric alone. \textcolor{black}{Although this formulation does not guarantee the preservation of all low-activity but important units, the empirical results in Table \ref{accuracy-1bit} demonstrate that combining these two factors improves pruning performance across the evaluated datasets.}
The effectiveness of the proposed importance score is experimentally validated in Section \ref{experiement:results}. Consequently, the pruning objective in Eqn. \ref{objective} becomes:
\begin{equation}
\arg \min_{h,n} \left( \sum_{i \in S(h)} I(h_i)+ \sum_{j \in S(n)} I(n_j) \right)
\end{equation}
After computing the per-unit importance scores, neurons and attention heads are ranked accordingly. The lowest-ranked units are then pruned to satisfy the ACs constraint. Following this initial pruning, we perform a mask search and refinement phase, where a greedy algorithm iteratively swaps pruned and unpruned units within each layer. This refinement step helps account for intra-layer dependencies and enhances the effectiveness of the pruning strategy. 

\subsubsection{Temporal Pruning} In the post-training stage, we also introduce a temporal pruning strategy to further improve efficiency in Spiking LLMs. In traditional SNNs, every timestep contributes to the forward pass, which can lead to inefficiencies, especially when many timesteps carry redundant or low-importance information. Moreover, not all layers may require the same temporal resolution. 
\textit{We hypothesize that more important layers in a network require more time unrolling. To address this, we propose a temporal pruning strategy that dynamically adjusts the number of timesteps per layer based on its temporal complexity.} To quantify this temporal complexity, we apply Principal Component Analysis to the layer-wise activations across timesteps. This approach is inspired by Ref. \cite{chakraborty2020constructing}, which employed PCA to infer the intrinsic dimensionality and redundancy of neural network layers in the context of adaptive quantization. Other works have used the PCA-based structural pruning technique in Refs. \cite{Garg_2020,chowdhury2021spatiotemporalpruningquantizationlowlatency}. Distinct from earlier approaches, our method targets the analysis of temporal dynamics rather than spatial redundancy. In our work, we record the ASR of each layer at every timestep, forming a temporal
activity matrix $\boldsymbol{\Phi} \in \mathbb{R}^{N \times T_{conv}}$, where $N$ denotes the total number of network layers (where each sublayer within the encoder blocks is considered separately) and $T_{conv}$ denotes the number of timesteps used to represent the temporal spiking dynamics. PCA is then applied to the temporal trajectories to determine the minimum number of principal components required to explain a target proportion of the
temporal variance. This effectively captures the complexity of its temporal dynamics.  
Subsequently, we map the PCA-derived importance of each layer to the number of timesteps allocated per layer using a nonlinear function to emphasize differences in temporal complexity across layers. The mapping function can be formulated as:
\begin{equation}
t^l = \left\lfloor \frac{b^{c^l}}{\max_j b^{c^j}} \cdot T_{conv} \right\rfloor
\end{equation}
where, $t^l$ denotes the number of timesteps allocated to layer $l$, $c^l$ is the PCA-derived importance score of layer $l$, $b$ is a scaling factor (base) for the power function, and $T_{conv}$ is the number of timesteps required for convergence. To ensure that layers with higher importance scores run for more timesteps, the scaling factor $b$ must be greater than 1. This mapping ensures that layers with higher temporal complexity are assigned more timesteps, preserving their dynamic representational capacity. Conversely, layers with lower temporal complexity receive fewer timesteps, allowing for more aggressive temporal pruning without significantly compromising performance. 
\begin{figure}[h]
    \centering
    \includegraphics[scale=0.75]{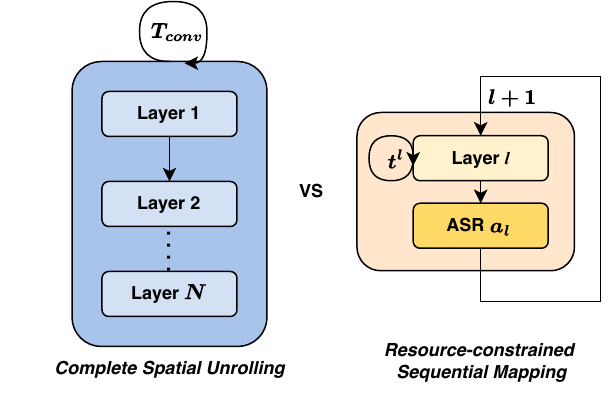}
    \caption{Illustration of two approaches for processing Spiking LLMs in hardware. (Left) The complete spatial unrolling structure, where spiking activity propagates through all $N$ layers with each layer processing the same number of timesteps $T_{conv}$. (Right) The resource-constrained sequential mapping structure, where each layer processes its assigned timesteps ($t^l$) independently, and the average spiking rate ($a_l$) is computed to guide the forward pass for the subsequent layer.}
    \label{fig:stochatic_llm}
\end{figure}

It is important to discuss the architectural dataflow in the context of our proposed temporal pruning. From a high-level hardware implementation perspective, there are two primary approaches to processing Spiking LLMs: (1) \textit{complete spatial unrolling}, where all layers are simultaneously mapped onto the chip, and (2) \textit{resource-constrained sequential mapping}, where layers are processed one at a time due to hardware limitations. The first approach allows parallel computation but requires significant hardware resources, which is often impractical for large-scale models. The second one is a more realistic approach that reuses limited compute resources by sequentially allocating them to different layers, as shown in Fig. \ref{fig:stochatic_llm}. The left panel depicts the first structure of Spiking LLMs, where spiking activity propagates sequentially through all layers, with each layer processing the same number of timesteps during the forward pass. 
The right panel shows the second approach, which facilitates the reuse of limited hardware resources by sequentially allocating them to different layers. This is relevant for our scenario to implement dynamic timestep allocation to various layers of the network. However, it necessitates storing the complete set of output spikes from all timesteps of a given layer for use by subsequent layers, which leads to significant memory overhead. To solve this problem, we propose stochastic spiking LLMs, which adopts a stochastic approach to generate the input spike trains to each layer. Instead of storing all the spikes at each timestep, which would be memory-intensive, we utilize the average spiking rate of the previous layer to stochastically generate spikes for the current layer. The equation for generating the input spikes in layer $l$ based on the ASR of the previous layer $a_{l-1}$ can be formulated as:
\begin{equation}
S_l^i \sim \textcolor{black}{\text{Bernoulli}(a_{l-1}^i)}
\end{equation}
where, $S_l^i \in \{0, 1\}^{t^l}$ is the binary spike train for neuron $i$ in layer $l$, with the length $t^l$ being the number of timesteps allocated for layer $l$. Each element of the spike train is independently sampled from a Bernoulli distribution with the probability $a_{l-1}^i$, the ASR of neuron $i$ in the previous layer $l-1$. By sampling from this Bernoulli distribution, we effectively simulate the spiking behavior without the need to retain detailed spike histories, thus saving memory and computational power. This is possible since we rely on rate encoding for our SNN operation. 


\subsection{Pruning During Retraining}\label{AA}
Spatial pruning and temporal pruning can also be incorporated effectively in the retraining stage to further enhance the compression efficiency. We leverage the head and neuron masks, as well as the layer-wise timestep allocations, obtained from the post-training pruning stage. In order to preserve spatial pruning while adhering to computational cost constraints, we incorporate a computational cost penalty into the loss function. Similar to the approach outlined in Ref. \cite{shi2024towards}, we modify the loss function to include a penalty term proportional to the number of ACs required by the pruned network. Since the number of ACs in SNNs depends on both spatial and temporal factors, the ACs should also consider the number of timesteps allocated to each layer during temporal pruning. The new loss function is given by the following equation:
\begin{equation}
\label{ltotal}
\mathcal{L}_{total} = \mathcal{L}_{pred}(h, n, t) + \lambda \cdot M(h, n, t)
\end{equation}
where, $\mathcal{L}_{pred}(h, n, t)$ is the standard prediction loss of the pruned model with head mask $h$ and neuron mask $n$. $M(h, n, t)$ represents the total number of ACs required by the pruned network, which is a function of the head mask $h$, neuron mask $n$, and timesteps list $t = [t^1, t^2, \ldots, t^l]$, where each element $t^l$ denotes the number of timesteps assigned to layer $l$. $\lambda$ is a hyperparameter controlling the trade-off between minimizing the loss and reducing the computational cost, effectively balancing performance and computational efficiency. Due to the binary nature of the mask, there is a non-differentiability problem \cite{shi2024towards}. To enable gradient-based training, we adopt a surrogate gradient where we approximate the binary mask using a sigmoid-based function. 

We employ a staged optimization strategy: During the initial training epochs, we incorporate the ACs penalty in the loss function to encourage low computational usage. Once the model stabilizes, we switch back to the original loss function to recover and preserve accuracy. This staged optimization helps balance efficiency with performance. Temporal pruning also utilizes post-training results as the initialization point. During fine-tuning, the model is first trained for several epochs to update its spiking activity. PCA analysis is then performed at regular intervals on the updated layer-wise spiking activity to reallocate the number of timesteps for each layer, instead of being applied at every epoch. This periodic update allows the model to adapt to its evolving temporal dynamics in a stable and efficient way.

\subsection{Additional Optimizations}

We augment our training framework with a few additional optimizations to improve the sparsity and energy efficiency of our Spiking LLMs.

The threshold potential is pivotal in spike generation and serves as a crucial hyperparameter in SNN design.  Unlike conventional approaches where a fixed threshold is applied uniformly across all layers, we treat the threshold $V_{th}$ in Eqn. \ref{eqn1} as a learnable parameter that is independently fine-tuned for each layer during training. Therefore, the network can allow each layer to adapt its spiking behavior.
This flexibility is especially important in deep spiking networks, where different layers may require different levels of sparsity and responsiveness.

Additionally, sparsity is another key characteristic of SNNs, contributing directly to their energy efficiency. While our pruning strategies already reduce computational cost, promoting sparsity offers an additional aspect of optimization. To explicitly encourage sparse spiking activity during training, we introduce a novel loss function, given as $
\mathcal{L}_{total} = \mathcal{L}_{pred}(h, n) + \lambda \cdot M(h, n, t)+\eta \cdot \mathcal{L}_{S}$. The hyperparameters $\lambda$ and $\eta$ control the trade-off between task performance, computational cost, and spiking sparsity. The activity loss function $\mathcal{L}_S$ is given as,
\begin{equation}
\begin{aligned}
\label{eqn4}
\mathcal{L}_{S}(a_1^*, \dots, a_L^*) = \sum_{l=1}^L{||a^*_l||_p}
\end{aligned}
\end{equation}
where, $L$ denotes the number of encoder layers, $||a^*_l||_p$ is the $p$-norm of the converged ASR for the output of the $l$-th spiking encoder layer. In our case, we use $p = 2$, corresponding to the Euclidean norm. By incorporating $\mathcal{L}_S$, the model achieves improved sparsity without compromising pruning effectiveness or task performance.

\section{Experiments}
\begin{table*}[h]
\centering
\caption{Hyperparameters Used During Pruning of SpikingBERT Models for Various Datasets}
\label{Hyperperameters-1bit}

\begin{tabular}{l|cccccccccc}
\hline

\textbf{Dataset} & \textbf{MNLI}&\textbf{QQP} & \textbf{QNLI} & \textbf{SST-2} & \textbf{MRPC}\\ \hline
$T_{conv}$ &140&80&110&85&110\\
Initial $V_{th}$ &0.85&1.0&0.9&1.0&0.9\\
Computational Cost Penalty $\lambda$ &1e-12&2e-12&2e-12&5e-12&5e-12\\
Training Batch Size &32&80&16&16&32\\
Testing Batch Size & 128 & 128&128&32&128\\
Learning Rate &5e-8&1e-7&1e-7&5e-10&1e-8\\ 
ACs Constraint&0.6&0.6&0.6&0.6&0.6 \\
PCA Scaling Factor $b$&1.02&1.02&1.02&1.02&1.02 \\
Activity Loss Scaling Factor $\eta$& 0.002&0.001&0.0015&0.001&0.0005\\
\hline

\end{tabular}
\end{table*}

\begin{table*}[h]
\centering
\caption{Performance Comparison of Various Pruning Methods on 1-Bit SpikingBERT Model Across Different Tasks}
\label{accuracy-1bit}

\begin{tabular}{lcccccccccc}
\hline

\multirow{3}{*}{\textbf{Dataset}} & \multicolumn{2}{c}{\textbf{MNLI}} & \multicolumn{2}{c}{\textbf{QQP}} & \multicolumn{2}{c}{\textbf{QNLI}} & \multicolumn{2}{c}{\textbf{SST-2}} & \multicolumn{2}{c}{\textbf{MRPC}} \\ 
& Acc & ACs & Acc & ACs& Acc& ACs & Acc & ACs & Acc& ACs \\ 
& \textbf{(\%)} & \textbf{(\%)} & \textbf{(\%)} & \textbf{(\%)} & \textbf{(\%)} & \textbf{(\%)} & \textbf{(\%)} & \textbf{(\%)} & \textbf{(\%)}& \textbf{(\%)}
\\
\hline

Baseline      & 75.54 & 100 & 83.99 & 100 & 80.85 & 100 & 86.58  & 100 & 75.71 & 100 \\ \hline
\multicolumn{11}{c}{\textbf{SPATIAL PRUNING}} \\ \hline

Post-Training (FIM)    & 72.03 &  60& 82.17 &  60& 80.27 & 60 & 85.32& 60 & 74.43 & 60\\ 
Post-Training (FIM + ASR of Neurons) & 72.77 & 60 & 82.63 & 60 & 80.67 & 60 & 85.67&60 & 74.67 & 60 \\ 
Post-Training (FIM + ASR of Attn) & 72.5 & 60 & 82.53 & 60 & 80.52 & 60 & 85.44&60 & 74.84 & 60 \\ 
Post-Training (FIM + ASR) & 73.17 & 60 & 82.84 & 60 & 80.82 & 60 &85.89&60&  75.01& 60\\ 
\textbf{Post-Training (FIM + ASR) + Retraining} & \textbf{74.97} & \textbf{54.17} & \textbf{84.24} & \textbf{57.95} & \textbf{81.88} & \textbf{39.96} & \textbf{86.47}&\textbf{60} & \textbf{76.12} & \textbf{33.18}\\ \hline

\multicolumn{11}{c}{\textbf{TEMPORAL PRUNING}} \\ \hline

Post-Training (PCA) & 75.26 & 56.05 & 82.63 & 64.22 & 77.16 & 62.87 & 86.47&61.68 & 74.61 & 57.96\\ 
\textbf{Post-Training (PCA) + Retraining} & \textbf{75.33}  & \textbf{56.05} & \textbf{84.01} & \textbf{68.66} & \textbf{81.93} & \textbf{62.87} & \textbf{86.70} & \textbf{47.14}& \textbf{75.71} & \textbf{59.72} \\ \hline
\multicolumn{11}{c}{\textbf{COMBINED SPATIAL AND TEMPORAL PRUNING}}  \\ \hline

\textbf{Two-Stage Pruning} & \textbf{75.13} & \textbf{41.67} & \textbf{84.10} & \textbf{41.82}   & \textbf{82.28} & \textbf{28.26} &\textbf{87.27} & \textbf{41.53} & \textbf{76.17}&\textbf{18.19}\\
\hline

\end{tabular}
\end{table*}


\subsection{Datasets}
We evaluated our model's performance using a diverse set of text classification tasks from the General Language Understanding Evaluation (GLUE) benchmark \cite{wang2019gluemultitaskbenchmarkanalysis}. Specifically, we employed the Quora Question Pair (QQP) and Microsoft Research Paraphrase Corpus (MRPC) datasets to assess the model's capability in handling similarity and paraphrase detection tasks. For inference-oriented evaluations, we utilized the Multi-Genre Natural Language Inference (MNLI) and Question-answering Natural Language Inference (QNLI) datasets to measure the model's ability to perform reasoning across various contexts and question-answering scenarios. For sentiment analysis, we leveraged the Stanford Sentiment Treebank (SST-2) dataset to evaluate the model's performance on single-sentence sentiment classification tasks. For all experiments, we set the maximum sequence length to 128 tokens.

\begin{table*}
\centering

\caption{Performance Improvement Due to Pruning and Adaptive Thresholding on 1-Bit SpikingBERT Models Across Different Tasks}
\label{energy_comparison}
\begin{tabular}{l|ccc|ccc|ccc}
\hline
\multirow{2}{*}{\textbf{Dataset}}  & \multicolumn{3}{c|}{\textbf{Baseline model }} & \multicolumn{3}{c|}{\textbf{Pruned Model}} & \multicolumn{3}{c}{\textbf{Adaptive Threshold}} \\
 & Accuracy & Normalized \#C &Latency& Accuracy & Normalized \#C & Latency & Accuracy & Normalized \#C & Latency \\ \hline
MNLI & 75.54 & 0.219 & 140 & 75.13 & 0.080 & 99.6 & 73.71 & 0.022 & 28.2 \\
QQP  & 83.99 & 0.266 & 80 & 84.10 & 0.103 & 55 & 82.77 & 0.027 & 15.20 \\
QNLI & 80.85 & 0.239 & 110 & 82.28& 0.064& 69.56 & 79.97& 0.017& 19.40\\ 
SST-2 & 86.58 & 0.241 & 85 & 87.27 & 0.100 & 53.08 & 86.58 & 0.023 & 13.04 \\ 
MRPC & 75.71 & 0.272 & 110 & 76.17 & 0.043 & 54 & 74.67 & 0.013 & 17.48\\ \hline

\end{tabular}

\end{table*}




\begin{table*}[h]
\centering
\caption{Performance Comparison of Other Compression Methods on BERT Across Different Datasets}
\label{tab:bert_pruning_glue}

\begin{tabular}{lcccccccc}
\toprule
\textbf{Model} & \textbf{Spiking} & \textbf{Layer} & \textbf{Compression} \textbf{Method} &
\textbf{MNLI} & \textbf{QQP} & \textbf{QNLI} & \textbf{SST-2} & \textbf{MRPC}  \\
\midrule
BERT \cite{devlin2019bertpretrainingdeepbidirectional}& \ding{55} &12& - & 84.6& 91.1& 90.5& 93.5 & 88.9 \\
TinyBERT \cite{jiao2020tinybertdistillingbertnatural} & \ding{55}& 4 & KD &82.5& 87.7 &87.7&92.6&86.4 \\
\midrule
PLATON \cite{zhang2022platon}& \ding{55}& 12 & Unstructured Spatial Pruning  & 82.0 & 90.2 & 88.9 & 90.5 & 84.3\\

Kwon et.al 2022 \cite{kwon2022fastposttrainingpruningframework} & \ding{55}& 12& Structured Spatial Pruning &82.51&90.35&90.06&92.49&85.27\\

EBERT \cite{liu2021ebert}& \ding{55} &12 & Structured Spatial pruning &  83.3 & 88.8 & 89.6 & 93.4 & - \\

CoFi \cite{xia2022structured} & \ding{55}& 12& Structured Spatial Pruning & 80.6 & 90.1 & 86.1  &90.6  & 82.6 \\

PGB \cite{lim2025pgb}& \ding{55} & 12 & Semi-Structured Spatial Pruning & -& 90.1& 86.4& 89.6&78.2 \\
\midrule
BPSN \cite{SU2024106630}   &\ding{51} & 4 &  - &73.0 & 83.2&82.8&86.0&83.3\\
SpikingBERT \cite{bal2024spikingbert} & \ding{51} & 4 & KD &  78.10 & 86.82 & 85.20 & 88.19 & 79.17 \\
4-bits EQ-SpikeLM \cite{11152565} &\ding{51} &12 & Structured Spatial Pruning + Quantization & 75.2& 83.1&83.6&86.5&83.1\\
1-bit SpikingBERT \cite{bal2024exploringextremequantizationspiking}&\ding{51} & 4 & Quantization & 75.4 & 83.8 & 80.5 & 86.7 & 75.8\\
\midrule
\textbf{Ours} &\ding{51} & \textbf{4} & \textbf{Spatio-Temporal Pruning + Quantization + KD} & \textbf{75.13} & \textbf{84.10}&\textbf{82.28}&\textbf{87.27}&\textbf{76.17}\\

\bottomrule
\end{tabular}

\end{table*}

\subsection{Experimental Setup}
The experiments were conducted on a system equipped with 8 NVIDIA RTX A5000 GPUs, each with 24 GB of memory. For each run, we employed PyTorch DataParallel using two GPUs. In our experiments, we use task-specific 1-bit SpikingBERT model with 4 encoder layers as the baseline for each individual task. All models are initialized from a full-precision SpikingBERT \cite{bal2024spikingbert}. We quantize the model weights to 1-bit following the approach in \cite{bal2024exploringextremequantizationspiking}. To mitigate any performance degradation caused by quantization, we apply intermediate-layer knowledge distillation using the full-precision teacher model, as described in \cite{bal2024exploringextremequantizationspiking}. Finally, we fine-tune the quantized model on the corresponding task-specific dataset to obtain the final model. \textcolor{black}{For PCA-based temporal pruning, we set the cumulative explained-variance threshold to 0.99999 for all datasets.} The remaining hyperparameter settings are listed in Table \ref{Hyperperameters-1bit}.

To quantify the event-driven operation of SNNs, we incorporated a Normalized Operations metric. In contrast to ANNs, where all neurons are active during each forward pass, SNNs only activate fan-out neurons when a spike occurs. Therefore, the number of effective operations in SNNs depend not just on the model architecture, but also on the spiking activity during inference, typically characterized by the average spiking rate. We adopt a normalized operations metric as used in \cite{10.3389/fnins.2020.00535,10766565,bal2024spikingbert}. The normalized computational factor is defined as:

\begin{equation}
\text{Normalized \#C} = \frac{\sum_{l=2}^{N-1} a_l \times \text{Layer ACs}^{l+1}}{\sum \text{Layer ACs}}
\label{eq:normalized_ops}
\end{equation}
where, 
$a_l$ is the average spiking rate of layer $l$, calculated as the total number of spikes generated by all neurons in the layer divided by the number of neurons and the number of timesteps.  $\text{Layer ACs}^{l+1}$ represents the number of AC operations in layer ${l+1}$.  In our experiments, we employ the Normalized \#C metric to assess the efficiency improvements achieved by our pruning strategy and compare the results against the baseline model as well as other optimization approaches.
\subsection{Results}
\label{experiement:results}
The accuracy (Acc) and \textcolor{black}{relative} ACs of the 1-bit quantized SpikingBERT model using various pruning methods are summarized in Table \ref{accuracy-1bit}. \textcolor{black}{The ACs (\%) represent the estimated accumulation operations relative to the corresponding baseline, which is set to 100\%.}  In our post-training spatial pruning approach, we evaluate different importance scores to determine which neurons and heads to prune. Post-Training, which uses only the FIM as the importance score, resulted in the lowest accuracy among the methods tested. To improve accuracy, Post-Training (FIM + ASR of Neurons) and Post-Training (FIM + ASR of Attn) consider the ASR of neurons and the attention output of each head in the importance scores, separately. The results show that they both can improve accuracy. Post-Training (FIM + ASR) combines these three scores, which achieves the highest accuracy. After obtaining the pruning masks of neurons and attention heads from the post-training spatial pruning stage, we apply a retraining pruning stage to further improve the model's performance. During retraining, we only train for a small number of epochs. Despite the short retraining period, this approach results in a notable improvement in accuracy. 

For temporal pruning, in the retraining stage, the model is fine-tuned with the newly constrained temporal structure. By allowing the network to adapt to the reduced timestep configuration, retraining helps the model to re-optimize its parameters and adjust its temporal processing. Two-stage pruning integrates spatial and temporal pruning. We use the retrained spatially pruned models to perform temporal pruning (post-training and retraining). Additionally, we also incorporate the sparsity improvement optimization strategy by including the activity loss in the loss function during this final step. 
To further evaluate the robustness of our method, we repeat the final retraining stage of the proposed two-stage pruning framework five times with different random seeds and report the mean and standard deviation of the resulting accuracy in Appendix \ref{five}.

Further, by incorporating the adaptive thresholding mechanism with our pruned model during the final retraining stage, the number of required timesteps can be further reduced. \textcolor{black}{This comes with a modest reduction in accuracy, reflecting a trade-off between computational efficiency and performance.} In our experiments, we used the pruned timestep list and scaled it down to a lower range. We then retrained the model using this new timestep list in combination with the adaptive threshold mechanism. The adaptive threshold allows layers to dynamically adjust their firing behavior in response to this scaling. The accuracy, latency, and computational cost of various 1-bit quantized models are presented in Table \ref{energy_comparison}. It is worth mentioning here that the latency is reported as the average number of timesteps across all layers. \textcolor{black}{The reported computational operations and average timesteps are model-level estimates rather than direct measurements of hardware energy consumption or inference latency.}  Compared with the baseline model, our pruning method significantly reduces both \textcolor{black}{estimated} computational cost and latency. Specifically, the total Normalized \#C metric is reduced by approximately 58.51\%-84.19\% compared to the baseline values, while the mean latency is reduced by 28.86\%-50.91\%. The adaptive thresholding mechanism can further reduce the latency and computational cost. The normalized \#C can be reduced by up to 89.85\%-95.22\%, while latency is reduced by 79.86\%-84.66\% relative to the baseline. These significant reductions in computational operations and timesteps can potentially translate into practical energy and latency benefits of neuromorphic processors, which can leverage event-driven benefits of sparse spiking events \cite{8259423}. 

Additionally, we compare our approach with a wide range of representative ANN- and SNN-based state-of-the-art compression methods on BERT, as summarized in Table \ref{tab:bert_pruning_glue}. Among ANN-based pruning methods, PLATON \cite{zhang2022platon} applies unstructured pruning to BERT-base, preserving only 10\% of the original weights. Kwon et al. \cite{kwon2022fastposttrainingpruningframework} and EBERT \cite{liu2021ebert} adopt the structured pruning approach, reducing the FLOPs of BERT-base by 59-60\%. CoFi \cite{xia2022structured} also employs structured pruning and achieves up to 95\% sparsity. PGB \cite{lim2025pgb} introduces a semi-structured pruning strategy with an 88\% pruning rate. For spiking LLMs, EQ-SpikeLM \cite{11152565} employs 4-bit quantization together with structured spatial pruning at a pruning ratio of 90\% using a 12-layer architecture. 1-bit SpikingBERT \cite{bal2024exploringextremequantizationspiking} further pushes quantization to the extreme by adopting 1-bit representations, extending SpikingBERT \cite{bal2024spikingbert} trained via knowledge distillation. \textcolor{black}{ Although these methods employ different model architectures and training settings, Table \ref{tab:bert_pruning_glue} provides a broader comparison of representative BERT compression approaches, highlighting the performance of our proposed spatio-temporal pruning framework on a compact 4-layer, 1-bit SpikingBERT.}

Unlike most compression methods developed for LLMs, spiking LLMs introduce an additional optimization dimension in the temporal domain. Our approach proposes a novel spatio-temporal pruning strategy that jointly reduces redundant spatial structures and temporal computation, while achieving high accuracy with low energy consumption and low inference latency under a more compact architecture. Further, our approach integrates spatio-temporal pruning, 1-bit quantization, and knowledge distillation, providing an effective compression framework for spiking LLMs.


\subsection{Ablation Studies}
\label{experiement:ablation}
\begin{figure}[htp]
    \subfigure[]{
    \includegraphics[scale=0.272]{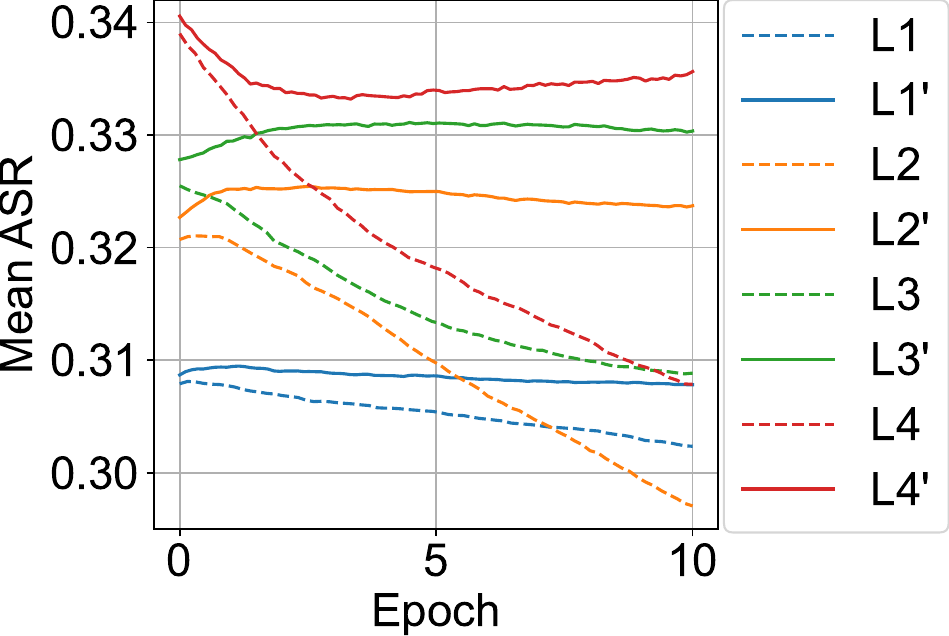}
    \label{asr_per_layer}}
    \hspace{-3mm}
    \subfigure[]{
    \includegraphics[scale=0.277]{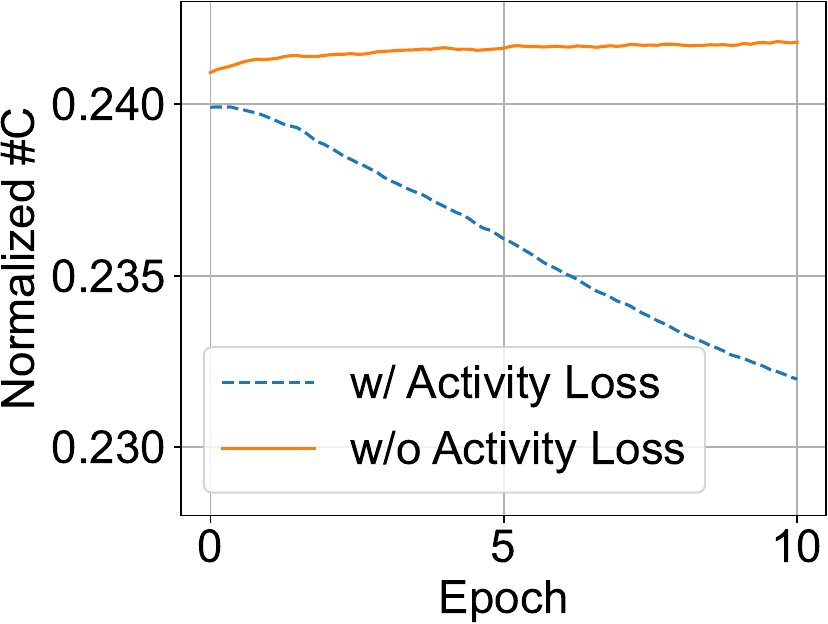}
    \label{activity}}
    \caption{Mean ASR per encoder layer and normalized number of computational operations (\#C) during training with and without activity loss for the 1-bit SpikingBERT model using the SST-2 dataset. (a) Solid lines (L1, L2, L3, L4) represent training with the activity loss applied to each encoder layer, while dashed lines (L1$'$, L2$'$, L3$'$, L4$'$) represent training without the activity loss. The activity loss effectively reduces mean ASR across all layers. (b) Training with the activity loss (blue dashed line) significantly reduces computational activity over epochs, highlighting its effectiveness in promoting energy-efficient sparse spiking behavior.}
\end{figure}

\noindent \textbf{Effect of Activity Loss:} We performed experiments to evaluate the impact of the activity loss term $\mathcal{L}_S$ on model behavior.
The steady-state behavior of the mean layer-specific ASR in the optimized model and the layer-wise reduction in neuronal activity due to activity loss $\mathcal{L}_S$, is shown in Fig. \ref{asr_per_layer} and \ref{activity}. We see that $\mathcal{L}_S$ leads to a notably lower ASR. This reduction indicates that the activity loss effectively suppresses redundant neuronal activity, leading to a more energy-efficient network.

\begin{figure}[t]
    \subfigure[]{
    \includegraphics[scale=0.283]{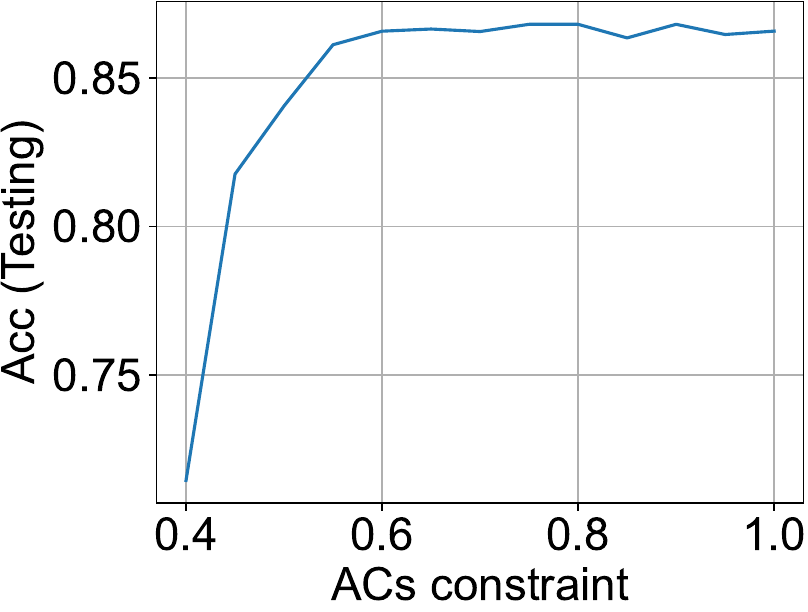}
    \label{spatial}}
    \hspace{-3mm}
    \subfigure[]{
    \includegraphics[scale=0.272]{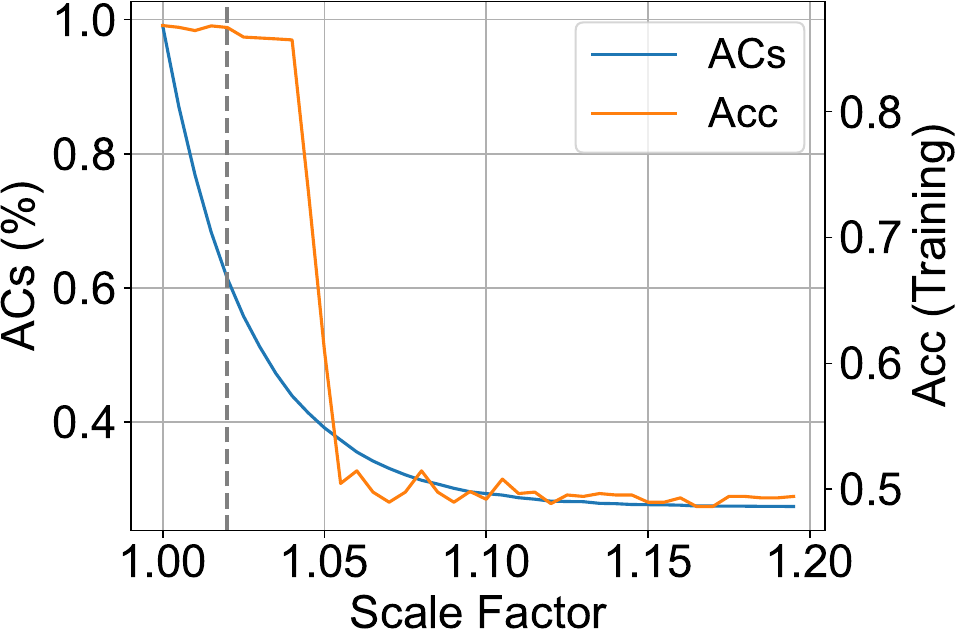}
    \label{scale}}
    
    \caption{Ablation studies on the effect of ACs constraint in spatial pruning and PCA scaling factor in temporal pruning, evaluated using the 1-bit quantized model on the SST-2 dataset. (a) The plot illustrates testing accuracy under varying ACs constraints during spatial pruning. (b) The plot shows the percentage of ACs retained relative to the baseline and training accuracy across different PCA scaling factors during temporal pruning. The dotted line shows the optimal value chosen for the PCA scaling factor.}
\end{figure}

\noindent \textbf{ACs Constraint Selection:} To evaluate the impact of computational constraints in post-training spatial pruning, we conducted experiments to explore various ACs budget settings. Specifically, we varied the ACs constraint from 0.4 to 1.0, which means that the pruned model is restricted to use only 40\% to 100\% of the original model’s ACs, as shown in Fig. \ref{spatial}. Based on the observed trade-off between model performance and computational efficiency, we selected the ACs constraint at 60\%. This setting achieves a desirable balance, significantly reducing the computational cost while maintaining acceptable performance. 

\noindent \textbf{PCA Scaling Factor Selection:} To determine the optimal scaling factor involved in mapping the PCA metric to the layerwise timestep allocation using a power function based non-linearity during temporal pruning, we performed experiments by varying the scaling factor from 1 to 1.2, and evaluating its impact on both the number of ACs and Acc of the model. As shown in Fig.~\ref{scale}, increasing the scaling factor progressively enforces stricter pruning, thereby reducing the total ACs. However, overly aggressive pruning leads to a significant drop in accuracy. Among the tested values, a scaling factor of 1.02 strikes an effective balance. It can preserve accuracy comparable to the baseline while achieving a notable reduction in ACs. 
\begin{figure}[t]
    \centering
    \begin{minipage}{0.5\columnwidth} 
        \centering       \includegraphics[width=\linewidth]{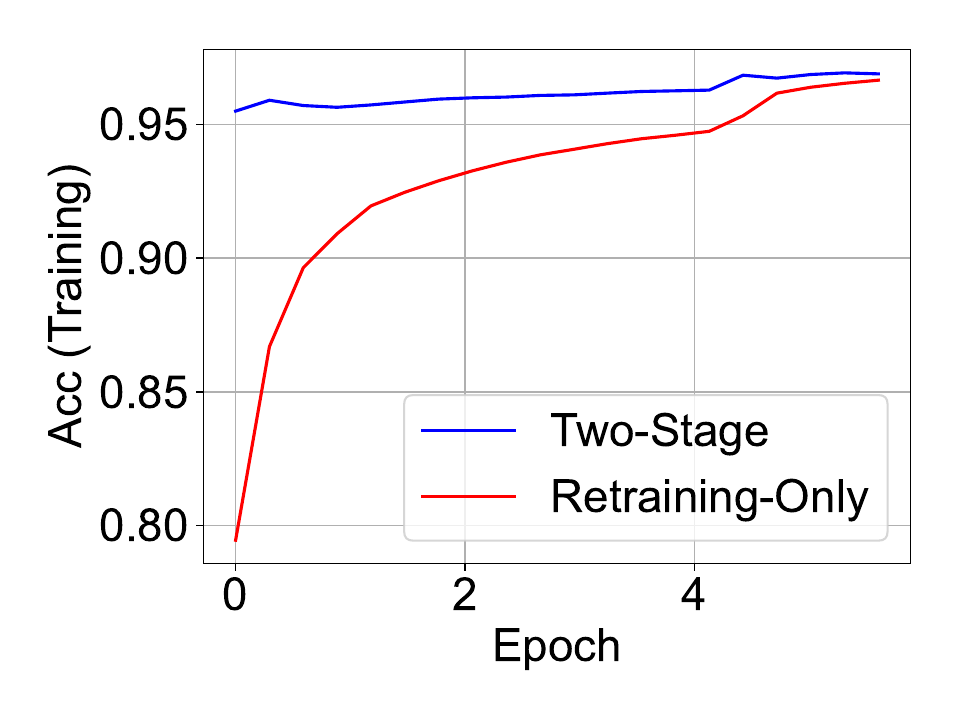}
        \label{fig:spatial}
    \end{minipage}%
    \hfill
    \begin{minipage}[t]{0.5\columnwidth}
    \centering
    \vspace{-5em}
    {\footnotesize  
    \begin{tabular}{l@{\hspace{0.15cm}}>{\centering\arraybackslash}p{1cm}@{\hspace{0.15cm}}>{\centering\arraybackslash}p{1.4cm}}
        \toprule
        \textbf{Dataset} & \textbf{Two-Stage} & \textbf{Retraining-Only} \\
        \midrule
        MNLI & 74.97 & 74.15 \\
        QQP & 84.24 & 83.73 \\
        QNLI & 81.88 & 81.02 \\
        SST-2 & 86.47 & 85.67 \\
        MRPC & 76.17 & 75.83 \\
        \bottomrule
    \end{tabular}
    } 
    \label{tab:comparison}
\end{minipage}
\vspace{-1.5em}
    \caption{Comparison of two-stage and retraining-only spatial pruning. (Left) Training accuracy and convergence curves on the SST-2 dataset using the 1-bit quantized model. The two-stage spatial pruning (blue) outperforms retraining from a random mask (red) in both accuracy and convergence speed. (Right) Testing accuracy across different datasets, comparing two-stage pruning with retraining-only approaches.}
    \label{fig:pruning_comparison}
\end{figure}

\noindent \textbf{Comparison of Two-Stage vs. Retraining-Only Spatial Pruning:} We evaluated the effectiveness of our two-stage spatial pruning strategy by comparing it with the retraining-only pruning, in which pruning masks are initialized randomly rather than being informed by a post-training phase. However, this approach did not achieve accuracies comparable to those obtained using the two-stage pruning. Furthermore, this approach required more epochs to train. As shown in Fig. \ref{fig:pruning_comparison}, the post-training followed by retraining approach achieved faster convergence and attains higher training accuracy compared to retraining from a random mask. The improved efficiency of the two-stage method is attributed to the initial pruning step, which removes less important neurons and attention heads, allowing subsequent retraining to more effectively fine-tune the model within the reduced architecture. 

\setlength{\tabcolsep}{4pt}
\begin{table}[h]
\centering
\caption{Comparison of Post-Training Pruning Strategies on SST-2 Dataset}
\begin{tabular}{lcccc}
\toprule
\textbf{Pruning Strategy} & \textbf{Spatial} & \textbf{Temporal} & \textbf{Retrain} & \textbf{Accuracy} \\
\midrule
No Pruning (Baseline)& \ding{55} & \ding{55} & 0 & 86.58 \\
Spatial Only& \ding{51} & \ding{55} & 1 & 86.47 \\
Temporal Only & \ding{55} & \ding{51} & 1 & 86.70 \\
Pruning + Joint Retraining& \ding{51} & \ding{51} & 1 & 85.55 \\
\textbf{Sequential} & \ding{51} & \ding{51} & \textbf{2} & \textbf{87.27} \\
\bottomrule
\end{tabular}
\label{tab:pruning_strategy_comparison}
\end{table}

\noindent \textbf{Two-Stage Spatial and Temporal Pruning Strategy:} We also designed experiments to investigate the optimal strategy for pruning our model. During post-training pruning, spatial and temporal pruning are applied independently to the baseline model. Once the best spatial masks and temporal timestep lists are identified, we proceed with retraining. Table \ref{tab:pruning_strategy_comparison} demonstrates that performing a sequential strategy that apply spatial pruning followed by retraining, and then applying temporal pruning with an additional retraining stage, leads to better accuracy compared to jointly retraining with both spatial and temporal pruning applied at once. Without intermediate recovery, the model struggles to cope with the simultaneous reduction in spatial capacity and temporal resolution, making convergence more difficult and increasing the risk of getting trapped in suboptimal local minima. Although our approach involves two retraining stages, the use of post-training pruning significantly reduces the number of epochs required for each retraining phase, making the overall process efficient.

\begin{table}[h]
\centering
\caption{Accuracy Comparison on SST-2 Dataset Between Baseline and Pruned Models Under Different Threshold Types and Timestep Settings}
\begin{tabular}{lccc}
\toprule
\textbf{Model} & \textbf{Timesteps} & \textbf{Accuracy (\%)} & \textbf{Threshold Type}\\
\midrule
Baseline & 85 & 86.58 & Fixed\\
Baseline & 21 &  82.80  & Adaptive\\
Pruned  & 53.08 & 87.27 & Fixed \\
Pruned & 13.04 & 86.58 & Adaptive \\
\bottomrule
\end{tabular}

\label{tab:pruned_vs_baseline}
\end{table}
\noindent \textbf{Effect of Adaptive Threshold:} To evaluate the impact of adaptive thresholding, we conducted experiments using the baseline model with the adaptive threshold mechanism. Compared to the baseline model without pruning, the adaptive threshold mechanism proves to be more effective when applied to a pruned model with reduced timesteps. As shown in Table. \ref{tab:pruned_vs_baseline},  we tested different thresholding strategies and timestep configurations using the 1-bit quantized model on the SST-2 dataset. It significantly improves performance in the pruned model under aggressive timestep reduction. This highlights the effectiveness of adaptive thresholding as a compensatory mechanism for pruning and reducing temporal resolution. Another contributing factor is our resource-constrained sequential mapping architecture, which introduces stochasticity during spike generation. This stochastic behavior avoids the typical spike propagation delay associated with sequential information propagation in deep spiking architectures, allowing it to maintain performance even when the number of timesteps is significantly reduced.

\begin{table}[H]
\centering
\color{black}
\caption{Accuracy Comparison Between Full Spike-Trajectory
Propagation and Stochastic Spike Propagation on GLUE}
\label{tab:spike_propagation}

\begin{tabular}{lccccc}
\hline
\textbf{Propagation Method}
& \textbf{MNLI}
& \textbf{QQP}
& \textbf{QNLI}
& \textbf{SST-2}
& \textbf{MRPC} \\
\hline
Full Spike-Trajectory
& 75.54 & 83.99 & 80.85 & 86.58 & 75.71 \\
Stochastic Bernoulli Sampling
& 74.88 & 81.10 & 78.44 & 86.12 & 73.97 \\
\hline
\end{tabular}
\end{table}
\noindent \textcolor{black}{\textbf{Effect of Stochastic Spike Propagation:} To evaluate the impact of stochastic spike propagation, we conduct a controlled comparison between full spike-trajectory propagation and Bernoulli sampling. Both approaches are evaluated using the same baseline model with identical weights and timesteps, without additional pruning or retraining. The only difference is the spike propagation scheme. The results are presented in Table \ref{tab:spike_propagation}. Compared to full spike-trajectory propagation, stochastic Bernoulli propagation introduces a modest accuracy reduction. This may be attributed to the loss of temporal correlations, as Bernoulli sampling preserves the average spiking rate but not the original temporal spike patterns. However, the results in Table \ref{accuracy-1bit} suggest that PCA-based temporal pruning followed by retraining can mitigate this accuracy degradation, even surpassing the original baseline accuracy on some datasets.}

\section{Conclusions}
This is the first work to jointly explore scalable spatial pruning, temporal pruning, extreme quantization, and knowledge distillation for Spiking LLMs evaluated on a large-scale NLP benchmark. By leveraging a two-stage pruning strategy, our method achieves substantial gains in latency and computational efficiency without compromising performance. These results demonstrate the effectiveness of combining SNNs with structured compression techniques to build highly efficient and deployable language models for energy-constrained applications. Future work can include extending the application of our pruning framework to other architectures and tasks, as well as exploring additional compression methods specifically tailored for spiking LLMs. By advancing these techniques, we aim to further bridge the gap between high-performance language models and energy-efficient computation on edge devices.

 \section*{Acknowledgments}
 This material is based upon work supported in part by the U.S. National Science Foundation under award No. CAREER \#2337646, CCSS \#2333881, CCF \#1955815, and EFRI BRAID \#2318101.


\vspace{12pt}
\appendices
\section{SpikingBERT Architecture}
\label{arch}
\begin{figure}[h]
    \centering
    \includegraphics[scale=0.61]{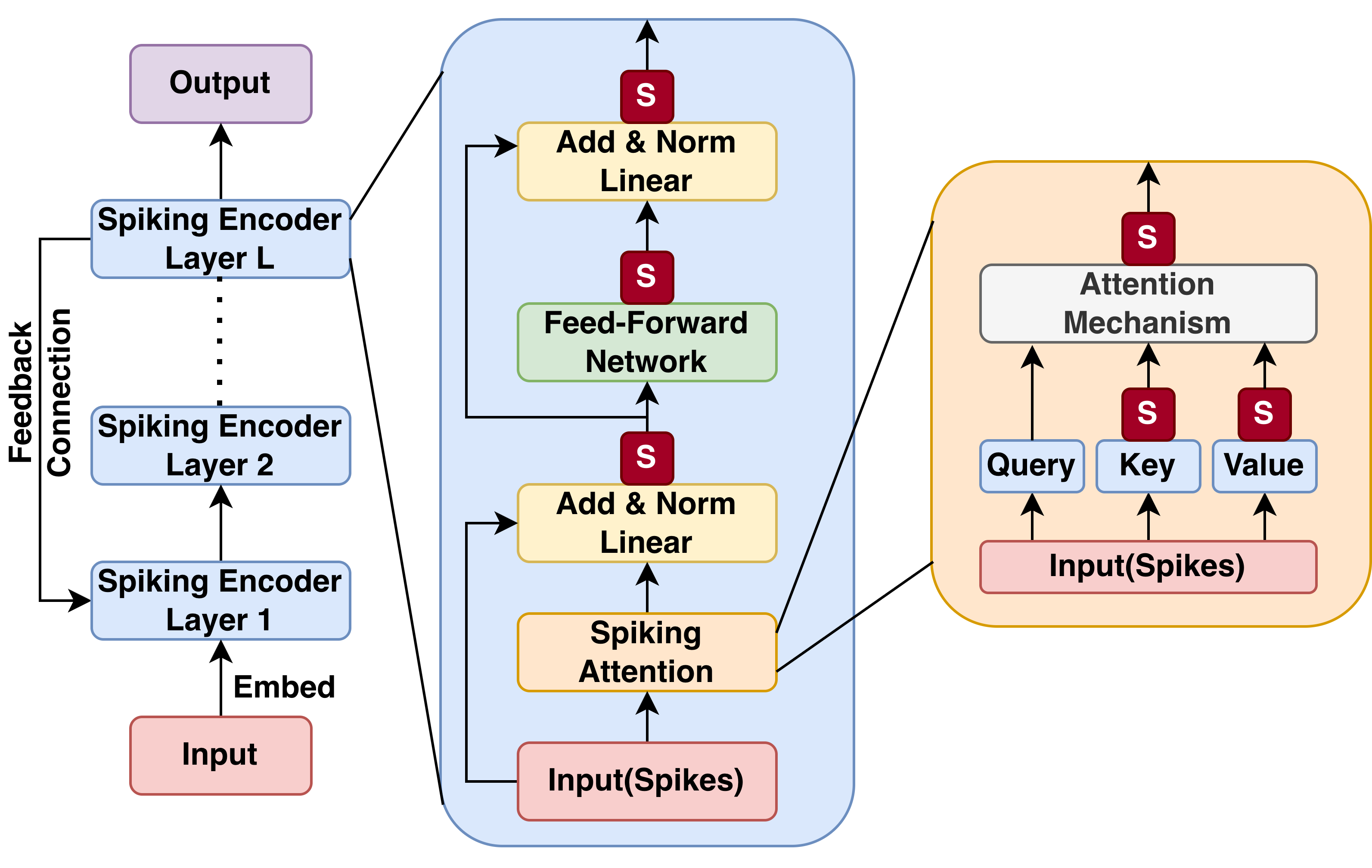}
    \caption{Overview of the SpikingBERT architecture. The model comprises of $L$ stacked Spiking Encoder layers. Each Spiking Encoder layer includes a Spiking Attention module followed by a series of spike-based layers. Spike generation ($S$) is governed by Leaky Integrate-and-Fire (LIF) neuron dynamics.}
    \label{fig:spikingBert}
\end{figure}
The architecture of the SpikingBERT model used in our work is shown in Fig.~\ref{fig:spikingBert}. This design retains the core structure of the Transformer encoder, with modifications to enable spike-based computation. In this architecture, each encoder block comprises of six spiking sublayers: (1) Key projection layer, (2) Value projection layer, (3) Spiking Attention layer, (4) the first Add \& Norm linear layer (self-attention output layer), (5) the Feed-Forward layer (intermediate layer), and (6) the second Add \& Norm linear layer (encoder output layer). The head mask is applied to sublayer (3), while the neuron mask is applied to sublayers (4) and (5). In our implementation, the Key and Value components of the attention mechanism are converted into spike-based representations, while the Query remains in its real-valued form (based on prior formulations \cite{bal2024spikingbert}). We therefore consider the Query projection and subsequent attention computation to be integrated within the spiking attention module. The attention output is subsequently converted into spikes for downstream processing. Additionally, we incorporated LIF neurons to generate spikes, along with residual connections and layer normalization modules, similar to the original BERT architecture.

\section{Computation of Number of Accumulation Operations}
\label{ACs}

We now detail the computation of the number of ACs, denoted as $M(h, n, t)$, as introduced in Eqn.~\ref{argmin} and Eqn.~\ref{ltotal}. Since each layer operates for a different number of timesteps, the equation can be expressed as:
\begin{equation}
M(h,n,t) = \sum_{l=1}^L{\text{ACs}_{h}^l+\text{ACs}_{n}^l}
\end{equation}
where, $L$ is the total number of encoder layers, $\text{ACs}_{h}^l$ represents the ACs for all attention heads in layer $l$, and $\text{ACs}_{n}^l$ represents the ACs for all neurons in layer $l$. $\text{ACs}_{h}^l$ is composed of three components: the ACs for the query, key, and value operations ($\text{ACs}_{\text{QKV}}^l$), the ACs for the attention computation ($\text{ACs}_{\text{Attn}}^l$), and the ACs for the fully connected self-attention output layer ($\text{ACs}_{\text{FC}}^l$). This relationship is summarized by the following equation:
\begin{equation}
\text{ACs}_{h}^l = \text{ACs}_{\text{QKV}}^l+ \text{ACs}_{\text{Attn}}^l + \text{ACs}_{\text{FC}}^l
\end{equation}
The ACs for the sum of the linear transformations of the query, key, and value operations can be represented as:
\begin{equation}
\text{ACs}_{\text{QKV}}^l = \text{ACs}_{\text{Q}} \times t_Q^l  + \text{ACs}_{\text{K}} \times t_K^l  + \text{ACs}_{\text{V}} \times t_V^l
\end{equation}
where, $\text{ACs}_{\text{Q}}$, $\text{ACs}_{\text{K}}$, $\text{ACs}_{\text{V}}$ are the ACs for the query, key and value respectively, scaled by their corresponding timesteps $t_Q^l$, $t_K^l$, and $t_V^l$. The ACs for the query, key, and value operations are equivalent and can be computed as:
\begin{equation}
\text{ACs}_{\text{Q}} = \text{ACs}_{\text{K}} = \text{ACs}_{\text{V}} = {N_{seq}} \times D\times H
\end{equation}
where, ${N_{seq}}$ is the sequence length, $D$ is the hidden size, and $H$ is the dimension of each attention head, which is equal to $D/{d^h}$. The ACs for the attention computation and for the fully connected self-attention output layer are given by the following equations:
\begin{equation}
\text{ACs}_{\text{Attn}}^l = 2 \times {{N_{seq}}}\times {{N_{seq}}} \times H \times t_\text{Attn}^l 
\end{equation}
\begin{equation}
\text{ACs}_{\text{FC}}^l = {{N_{seq}}}\times D \times H\times t_\text{FC}^l 
\end{equation}
where, $t_\text{Attn}^l$ and $t_\text{FC}^l$ denote the timesteps of the self-attention layer and self-attention output layer, respectively. In our architecture, since the Query projection is implemented within the spiking attention layer, we set $t_Q^l = t_\text{Attn}^l$. The attention layer contains two dot product operations for each head in layer $l$, first computing attention scores by comparing queries with keys, and the second for generating the weighted output by applying the attention scores to the values.

The ACs for all neurons in the $l^{th}$ layer are given by:
\begin{equation}
\text{ACs}_{n}^l = N_{seq} \times D \times t_\text{Inter}^l + N_{seq} \times D \times t_\text{Output}^l
\end{equation}
where $t_\text{Inter}^l$ and $t_\text{Output}^l$ denote the number of timesteps allocated to the intermediate and output layers, respectively.

Post-training spatial pruning requires a fixed architectural constraint to guide the pruning of attention heads and neurons. Since the pruning masks are determined after training without further tuning of spiking behavior, incorporating ASR introduces variability and undermines the determinism of the pruning process. Therefore, to maintain consistency and control over the computational budget during post-training pruning, we rely solely on structural metrics (e.g., number of heads or neurons) rather than dynamic activity-based measures like ASR.



\section{Statistical Results Across Multiple Runs}
\label{five}

To evaluate the stability of the final retraining stage, we repeat this stage five times using different random seeds, while keeping the preceding pruning stages fixed. Table \ref{tab:mean_std} reports the mean and standard deviation of the accuracy across five GLUE datasets. The relatively small standard deviations demonstrate the stability of the final retraining stage across the evaluated datasets.

\begin{table}[H]
\centering
\caption{Accuracy of the Proposed Two-Stage Pruning
Method Across Five Final-Stage Retraining Runs}
\label{tab:mean_std}

\begin{tabular}{lc}
\hline
\textbf{Dataset} & \textbf{Accuracy (\%)} \\
\hline
MNLI  & $75.22 \pm 0.16$ \\
QQP   & $83.93 \pm 0.14$ \\
QNLI  & $82.14 \pm 0.13$ \\
SST-2 & $86.72 \pm 0.37$ \\
MRPC  & $75.94 \pm 0.26$ \\
\hline
\end{tabular}
\end{table}

\textcolor{black}{
To further investigate the robustness of the pruning pipeline, we conduct five runs on the SST-2 dataset using different random seeds, starting from the same trained baseline model. We evaluate representative spatial and temporal pruning configurations, including their retraining stages. Table \ref{tab:pipeline_std} reports the mean and standard deviation of the accuracy across these runs, with the baseline accuracy included as a reference. The results show consistent accuracy across different random seeds for the evaluated pruning configurations.
}

\begin{table}[H]
\centering
\caption{Accuracy Comparison Across Five Pruning Runs
Using a Fixed Baseline on SST-2}
\label{tab:pipeline_std}

\begin{tabular}{lc}
\hline
\textbf{Method} & \textbf{Accuracy (\%)} \\
\hline
Baseline (1-bit SpikingBERT) & 86.58 \\
Post-Training (FIM + ASR) & $85.62 \pm 0.24$ \\
Post-Training (FIM + ASR) + Retraining
& $86.03 \pm 0.28$ \\
Post-Training (PCA) & $86.33 \pm 0.15$ \\
Post-Training (PCA) + Retraining
& $86.54 \pm 0.19$ \\
Two-Stage Pruning & $86.74 \pm 0.33$ \\
\hline
\end{tabular}
\end{table}

\end{document}